\pdfoutput=1
\documentclass[11pt]{article}

\usepackage[final]{acl}
\usepackage{times}
\usepackage{latexsym}
\usepackage{booktabs}
\usepackage{pifont}
\newcommand{\xmark}{\ding{55}}%
\usepackage{bm}
\usepackage{tikz}
\usepackage{graphicx}
\usepackage{multirow}
\usepackage{adjustbox}
\usepackage{amsmath}
\usepackage{comment}
\usepackage{amssymb}
\usepackage[T1]{fontenc}
\usepackage[utf8]{inputenc}

\usepackage{microtype}

\usepackage{inconsolata}

\usepackage{tcolorbox}

 {\begin{list}{}%
         {\setlength{\leftmargin}{#1}}%
         \item[]%
 }
 {\end{list}}

\title{Domain Expansion: Parameter-Efficient Modules as \\ Building Blocks for Composite Domains}

\author{Mann Patel, Divyajyoti Panda, Hilay Mehta, Parth Patel,  Dhruv Parikh}

\begin{document}

\maketitle

\begin{abstract}
Parameter-Efficient Fine-Tuning (PEFT) is an efficient alternative to full scale fine-tuning, gaining popularity recently. With pre-trained model sizes growing exponentially, PEFT can be effectively utilized to fine-tune compact modules, Parameter-Efficient Modules (PEMs), trained to be domain experts over diverse domains. In this project, we explore composing such individually fine-tuned PEMs for distribution generalization over the composite domain. To compose PEMs, simple composing functions are used that operate purely on the weight space of the individually fine-tuned PEMs, without requiring any additional fine-tuning. The proposed method is applied to the task of representing the 16 Myers-Briggs Type Indicator (MBTI) composite personalities via 4 building block dichotomies, comprising of 8 individual traits which can be merged (composed) to yield a unique personality. We evaluate the individual trait PEMs and the composed personality PEMs via an online MBTI personality quiz questionnaire, validating the efficacy of PEFT to fine-tune PEMs and merging PEMs without further fine-tuning for domain composition. Code is available here.
\footnote{\url{https://github.com/manncodes/domain-expansion}}
\end{abstract}

\section{Introduction}
\label{sec:intro}
Parameter-Efficient Fine-Tuning (PEFT) based methods freeze most of the parameters associated with pre-trained language models (PLMs) and fine-tune only a small sub-set of parameters (adapter parameters) to customize PLMs for downstream tasks. This allows for efficient and quick fine-tuning via PEFT to yield Parameter-Efficient Modules (PEMs), with a reduced model size and memory footprint \cite{houlsby2019parameterefficient}, trained on domain-specific datasets to be domain experts. Due to the exponentially increasing model sizes associated with PLMs \cite{zhao2023survey}, PEFT methods have become increasingly popular \cite{xu2023parameterefficient} and the default standard to fine-tune such PLMs. In this project, we choose LoRA \cite{hu2021lora} and IA3 \cite{liu2022fewshot} state-of-the-art PEFT approaches as our PEM architectures. 

\begin{figure}
\centerline{\includegraphics[width = \linewidth]{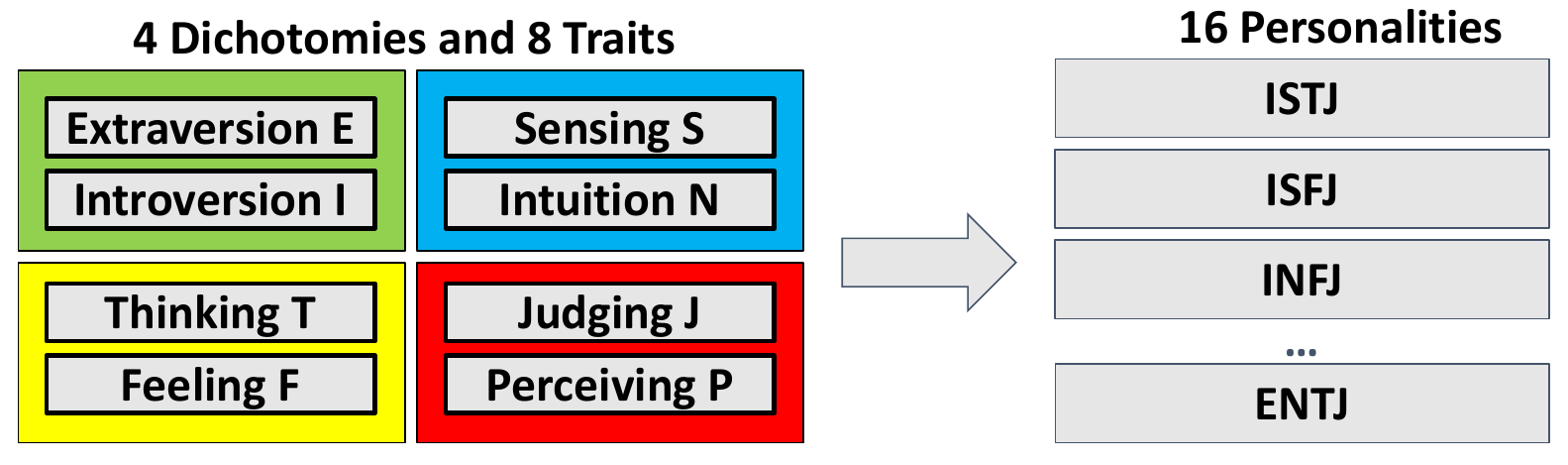}}
\caption{MBTI Dichotomies, Traits and Personalities}
\label{fig:mbti_exmpl}
\end{figure}

Myers-Briggs Type Indicator (MBTI) is a widely used framework to characterize personality traits of individuals \cite{mbtipaper}. An MBTI personality can be decomposed into 4 key traits, each trait associated with a dichotomy. Specifically, the 4 dichotomies in MBTI are: (i) Extraversion (\textbf{E}) - Introversion (\textbf{I}) (ii) Sensing (\textbf{S}) - Intuition (\textbf{N}) (iii) Thinking (\textbf{T}) - Feeling (\textbf{F}) (iv) Judging (\textbf{J}) - Perceiving (\textbf{P}). A single MBTI personality, such as \textbf{INTJ}, is constructed by composing 4 traits, each trait selected from the two opposite traits in each dichotomy. Fig. \ref{fig:mbti_exmpl} shows the composition of traits from dichotomies into the 16 personalities in MBTI.

In this project, we propose a novel method to represent each of the 16 MBTI personalities using language models. The problem of representing an MBTI personality, via a language model, is formulated as a PEM composition problem. In particular, as the MBTI personalities can be decomposed into their 4 characteristic traits, we first train a PEM for each of these traits, yielding a total of 8 trait PEMs (4 dichotomies $\times$ 2 opposing traits per dichotomy). Next, we compose these individual PEMs via a simple composition function, $f(.)$, operating purely on the weight (parameter) space (of the individual PEMs) and without any additional fine-tuning, to yield a personality PEM for each of the 16 MBTI personalities. We utilize the MBTI personality quiz questionnaire\footnote{https://www.16personalities.com/free-personality-test} to evaluate both, the individual trait PEMs and the composed personality PEMs.

\section{Related Work}
Several prior works discuss language model composition for distribution generalization to composite domain tasks. Specifically,  \cite{wortsman2022model, matena2022merging, jin2023dataless, ilharco2023editing}, perform a full fine-tuning of PLMs across tasks and/or domains (each PLM initialized from the same pre-trained checkpoint) and merge/compose the fine-tuned models by performing simple arithmetic operations on model parameters, improving the composed model performance. On the other hand, \cite{pfeiffer2021adapterfusion, wang2022adamix}, compose PEMs by fusing individual PEM outputs via a learnable module or via mixture-of-experts, both requiring post-composition fine-tuning. We utilize the method proposed by \cite{zhang2023composing} to compose PEMs \textit{without} requiring additional fine-tuning, applying it to the problem of representing MBTI personalities through PEMs.

Prior works that combine language models and MBTI personalities typically explore language models to classify the MBTI personality associated with an excerpt of text. \cite{dos_Santos_2022, keh2019myersbriggs, bottomtopmbti, xmer_mbti_pred} are works that fine-tune PLMs in order to classify text data with an MBTI personality. \cite{fernau22_interspeech} aligns a chatbot to the MBTI personality of its user in order to improve chatbot usability. 

However, no prior work uses PEMs as building blocks to compose personalities via PEM traits, fine-tuning language models (PEMs) to behave as individuals with trait/personality characteristics. Further, the proposed method is general, simple and efficient: PEMs are inexpensive to train, result in smaller model sizes and require no additional fine-tuning data post-composition for distribution generalization.

\section{Problem Description}
\label{sec:prob_des}
To formally define our problem, we consider $N$ domains, each domain representing an individual task and/or application. Given these $N$ domains, $D_1, D_2, ..., D_N$ we fine-tune a PEM for each such domain, yielding $N$ PEMs, $M_{\bm{\theta}_1}, M_{\bm{\theta}_2}, ..., M_{\bm{\theta}_N}$, where $\bm{\theta}_i$ indicates the (adapter) parameters of the $i^{th}$ PEM model, $M_{\bm{\theta}_i}$. We wish to compose such $N$ fine-tuned PEMs via a composing function $f(.)$ applied to the (adapter) parameters of the individual PEMs, to yield a composite PEM $\mathcal{M}_{\bm{\theta}_C}$ with composite (adapter) parameters $\bm{\theta}_C$ that generalizes over the composite task $D_C = \bigcup\limits_{i=1}^{N} D_i$.

\begin{align}
\bm{\theta}_C &= f(\bm{\theta}_1, \bm{\theta}_2, ..., \bm{\theta}_N) \\
D_C &= \bigcup\limits_{i=1}^{N} D_i 
\label{eq:eq_pem_comp}
\end{align}

Note that $\bm{\theta}_i$ are adapter parameters representing a small fraction of the PLM model parameters utilized for (parameter-efficient) fine-tuning (PEFT). Additionally, the fine-tuning (PEFT) performed for each $D_i$ is initiated from an identical PLM model checkpoint, across all the domains over which composition for distribution generalization is to be performed. 

\section{Proposed Method}

\begin{figure}
\centerline{\includegraphics[width = \linewidth]{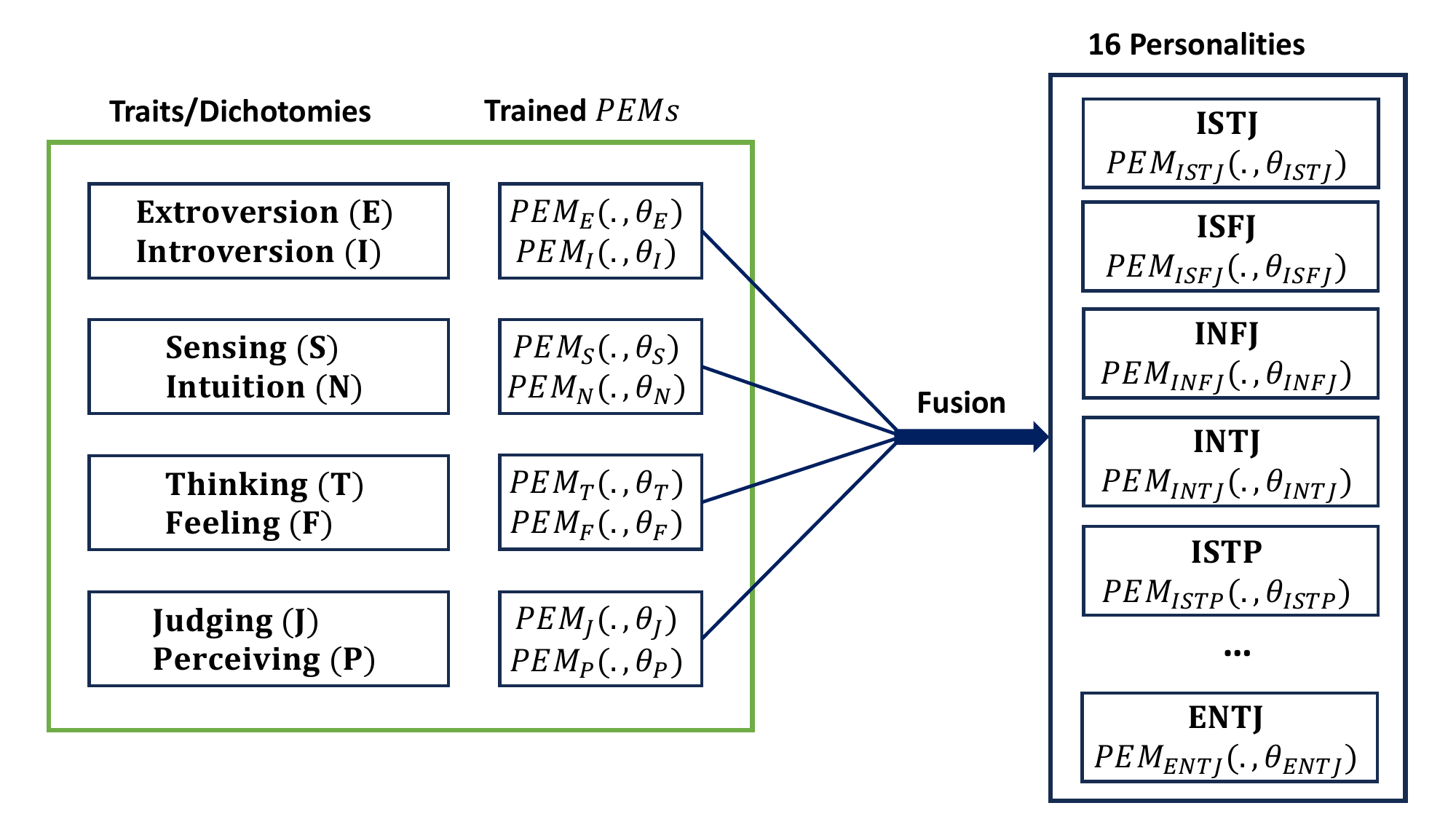}}
\caption{Method Overview}
\label{fig:method_ovw}
\end{figure}

In our project, the individual domains $D_i$ (section \ref{sec:prob_des}) correspond to the individual traits \textbf{E, I, S, N, T, F, J, P} defined in
section \ref{sec:intro} (2 opposing traits for each of the 4 dichotomies). The individual trait PEMs are first trained for each trait (domain) to yield 8 base PEMs. Next, we merge/compose 4 trait PEMs (selecting exactly one trait from each dichotomy) to yield a personality PEM (for a total of $2^{4} = 16$ personality PEMs). The overview of our proposed method can be seen in Fig. \ref{fig:method_ovw}.

\subsection{Trait PEMs via PEFT}
\begin{figure}
\centerline{\includegraphics[width = \linewidth]{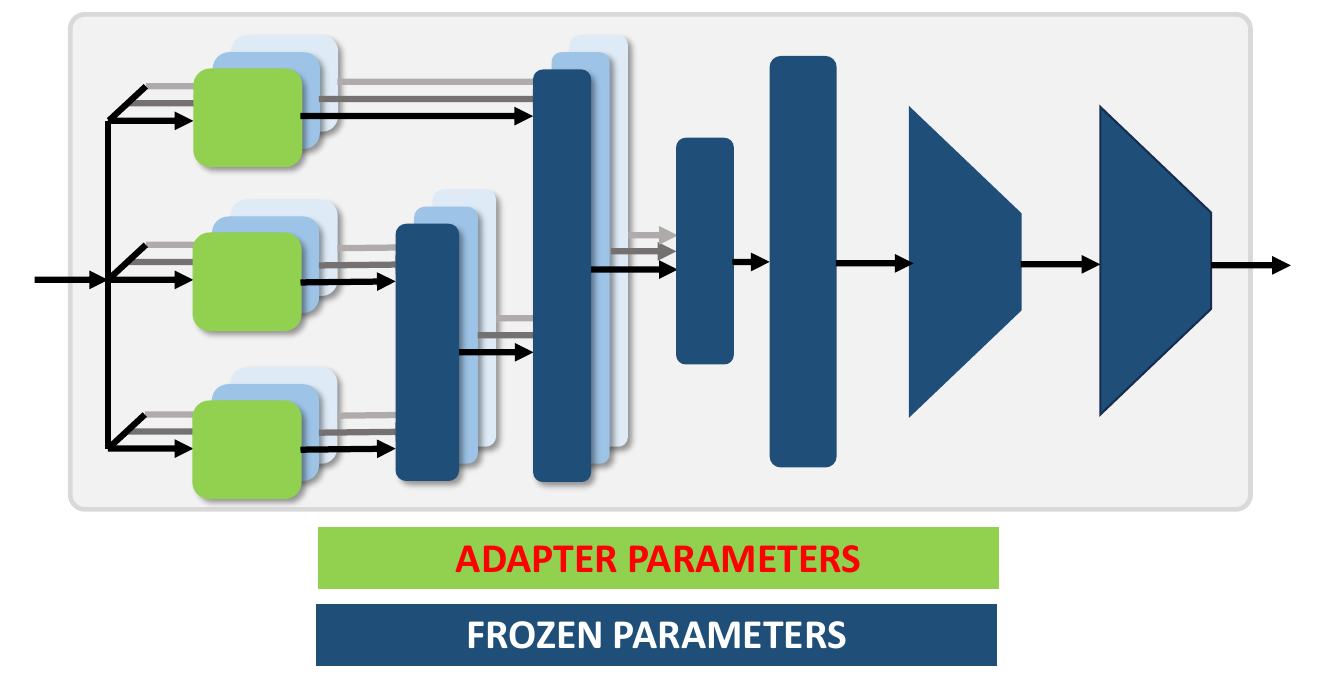}}
\caption{PEFT Training}
\label{fig:peft_ft}
\end{figure}

In order to train the individual trait PEMs, we utilize 2 state-of-the-art PEFT methods/architectures - (i) LoRA (ii) IA3. PEFT based approaches, including LoRA and IA3, work by freezing parameters associated with the base PLM excluding certain fine-tuning parameters (adapter parameters or adapters) (Fig. \ref{fig:peft_ft}). During the backward pass, the frozen parameters stay fixed (unmodified) while the adapter parameters are allowed to learn (update) from the domain data. PEFT, thus, (i) reduces effective model size (ii) makes training inexpensive/quick (iii) reduces the number of domain data samples required for fine-tuning (iv) can be used to train several diverse domain expert PEMs, efficiently. 

\paragraph{LoRA} LoRA \cite{hu2021lora} (Low-Rank Adaptation) is a prevalent PEFT method. For layers in a transformer that transform an input $\bm{x} \in \mathbb{R}^{k}$ to $\bm{h} \in \mathbb{R}^{d}$ via weight matrices, LoRA modifies this transformation as below,
\begin{align}
\label{eq:lora_1}
    \bm{h} \leftarrow \bm{h} + \bm{BAx} \\
    \bm{\theta}_{LoRA} = \{\bm{B}, \bm{A}\} 
\label{eq:lora_2}
\end{align}
In eq. \ref{eq:lora_1}, $\bm{B} \in \mathbb{R}^{d \times r}$ and $\bm{A} \in \mathbb{R}^{r \times k}$ are the LoRA adapter parameters $\bm{\theta}_{LoRA}$ with $r \ll \min(d, k)$. While the modification in eq. \ref{eq:lora_1} can be done for any $\bm{x} \rightarrow \bm{h}$ transformation within a transformer, it is conventionally applied for the query and value vector generation in each transformer layer. In our project, we follow this convention.

\paragraph{IA3} IA3 \cite{liu2022fewshot} is a PEFT method which was proposed for few-shot learning. In order to perform PEFT, IA3 introduces learnable scale vectors $\bm{l}_k$, $\bm{l}_v$ and $\bm{l}_{ff}$. A learnable scale vector $\bm{l} \in \mathbb{R}^{d}$ is applied to a vector $\bm{h} \in \mathbb{R}^{d}$ as below,

\begin{equation}
    \bm{h} \leftarrow \bm{h} \odot \bm{l}
\label{eq:ia3}
\end{equation}

In eq. \ref{eq:ia3}, $\bm{h}$ is a vector generated within the transformer while $\bm{l}$ is the adapter parameter for IA3. Particularly, $\bm{l}_k$ and $\bm{l}_v$ are applied to key and value vectors, respectively, in the multi-headed self-attention mechanism and $\bm{l}_{ff}$ is applied to the intermediate activations of the feed-forward network, in a transformer (in each layer). Thus, $\bm{\theta}_{IA3} = \{\bm{l}_k, \bm{l}_v, \bm{l}_{ff}\}$.

\subsection{Personality PEMs via PEM Composition}

Using LoRA and IA3, we train 8 trait PEMs (each, for both LoRA and IA3). To obtain a personality PEM, we compose (merge) the trait PEMs, without further fine-tuning post-composition. Given trait PEMs $M_{\bm{\theta}_{T_{i}}}$ where $T_{i}$ is the $i^{th}$ dichotomy trait, $i \in \{1, 2, 3, 4\}$ for the 4 dichotomies, we compose PEMs as below,

\begin{equation}
    \bm{\Theta}_{P} = \sum\limits_{i=1}^{4} \bm{\theta}_{T_{i}}
\label{eq:pem_composition}
\end{equation}

In eq. \ref{eq:pem_composition}, $P = T_{1}T_{2}T_{3}T_{4}$, is an MBTI personality type and $\bm{\Theta}_{P}$ are the adapter parameters obtained for the composite PEM $\mathcal{M}_{\bm{\Theta}_{P}}$ representing that personality. Note that the resulting composite PEM has the same architecture as the individual combining PEMs. The adapter parameters of the composite PEM is given by a simple sum (composing function $f \rightarrow \sum$) over the adapter parameters of the individual PEMs, leading to distribution generalization over the combining (trait) domains.

Fundamental works such as word2vec \cite{mikolov2013efficient} motivate such simple PEM composition rule for distribution generalization. Further, \cite{matena2022merging, wortsman2022model, jin2023dataless} hypothesize that models which are fine-tuned from the same initial PLM checkpoint often lie within similar error bins. Thus, the parameters of such models can simply be added up to yield a superior model with improved generalization.

\section{Experimental Results}
We worked on two tasks, zero-shot classification for classifying the prompt into one of the seven categories on the 7-point Likert scale (strongly agree, agree, slightly agree, neutral, slightly disagree, disagree and strongly disagree), and text generation which acted as a chatbot for answering questions.

For zero-shot classification, we utilize the BERT base model as the PLM backbone. In order to fine-tune PEM adapters for each individual trait, we insert a classification head (adapter parameter) on top of a BERT base model. For text classification, we use the RoBERTa model as the base model.

A PEM is fine-tuned for a trait such that it responds to input questions asked to it as if an individual with that trait would respond. Similarly, post-composition, the composite PEM responds to input questions in alignment with the resultant personality obtained from the composing (combining) trait (PEMs). 

\paragraph{Synthetic Dataset Generation} In order to train the individual trait PEMs, we synthetically generate 8 trait datasets (one per trait) by prompting ChatGPT-4. A generated data sample, from ChatGPT-4, for a given trait dataset, is in the format \{`question': <\textit{question}>, `answer': <\textit{answer}>\}. A <\textit{question}> corresponds to a question (statement)\footnote{Note that the question can also just be a simple statement to be classified} relevant to the given trait (dataset). For zero-shot classification, <\textit{answer}> is the response in the 7 classes that aligns with the trait. An example sample for the \textbf{Introversion} trait dataset is shown below,

\begin{tcolorbox}[colback=green!10!white,colframe=green!50!black]
\emph{\textbf{question}}: I find it energizing to be around other people for long periods of time. \\ 
\emph{\textbf{answer}}: Strongly Disagree
\end{tcolorbox}

For text generation, <\textit{answer}> is the response to the question provided that aligns with the trait. An example sample for the \textbf{Introversion} trait dataset is shown below,

\begin{tcolorbox}[colback=green!10!white,colframe=green!50!black]
\emph{\textbf{question}}: Describe a perfect weekend for you. \\ 
\emph{\textbf{answer}}: A perfect weekend for me would involve spending time alone or with one or two close friends, perhaps exploring a new book or diving into a personal project.
\end{tcolorbox}

We generate 154 samples per trait dataset. ChatGPT-4 is prompted \footnote{Not the exact prompt. We add a random seed and a temperature parameter to the prompt to ensure that samples are not repeated} as below for zero-shot classification and text generation respectively, as an example, to generate trait dataset for the trait \textbf{Judging},

\begin{tcolorbox}[colback=green!10!white,colframe=green!50!black]
\emph{\textbf{prompt}}: Draft me 154 question with answers being in the 7 classes which can be used to distinguish between \textbf{Judger} and \textbf{Perceiver} MBTI traits. Save the questions in a JSON format, responding as a \textbf{Judger}. Generate examples such that there are at least 5 samples for each of the 7 classes.
\end{tcolorbox}

\begin{tcolorbox}[colback=green!10!white,colframe=green!50!black]
\emph{\textbf{prompt}}: Draft me 154 question with answers which can be used to check whether the person is a \textbf{Judger} or a \textbf{Perceiver}. Save the questions in a JSON format, responding as a \textbf{Judger}.
\end{tcolorbox}

\paragraph{PEM Composition} Instead of using eq. \ref{eq:pem_composition} directly to compose trait PEMs, we use the below,

\begin{equation}
    \bm{\Theta}_{P} =  \sum\limits_{i=1}^{4} \lambda_i\bm{\theta}_{T_{i}}
\label{eq:pem_comp_act}
\end{equation}

In eq. \ref{eq:pem_comp_act}, the hyper-parameter set \{$\lambda_i$\} is selected such that each $\lambda_i \in (0, 1)$ and $\sum\limits_{i} \lambda_i = 1$. An optimal set of \{$\lambda_{i}^{*}$\} is obtained by sweeping over a set of possible values (following the constraints, with a granularity of 0.1) and selecting the \{$\lambda_i$\} maximizing the (personality) evaluation score.  

\paragraph{Evaluation} 
We evaluate the fine-tuned individual trait PEMs and the composed personality PEMs using the personality quiz questionnaire at https://www.16personalities.com/free-personality-test. The questionnaire comprises of 60 questions that a user responds to (with a response in one of the 7 classes) to identify their personality type. The questionnaire outputs the users personality type based on the responses, along with a percentage score for each trait (per dichotomy).

The questionnaire is used as input to the PEMs and the output responses are recorded and fed to the online questionnaire via an automated Selenium pipeline to automate evaluation.

\paragraph{Results}

\begin{table}
\centering
\begin{adjustbox}{width=0.5\textwidth}
\begin{tabular}{c|c|c|c}
\toprule
Trait        & Baseline (\%) & LoRA (\%) & IA3 (\%) \\ \midrule
Extroversion & 45            & 70        & 57       \\
Introversion & 55            & 72        & 72       \\
Intuitive    & 42            & 66        & 68       \\
Sensor       & 58            & 80        & 80       \\
Thinker      & 54            & 64        & 61       \\
Feeler       & 46            & 57        & 56       \\
Judger       & 49            & 64        & 64       \\
Perceiver    & 51            & 65        & 61       \\ \bottomrule
\end{tabular}
\end{adjustbox}
\caption{MBTI Trait Results}
\label{tab:trait}
\end{table}

% Trait & E/I & S/N & T/F & J/P & \checkmark/\xmark & E/I & S/N & T/F & J/P & \checkmark/\xmark \\ \midrule
% 
\begin{table}
\centering
\begin{adjustbox}{width=0.5\textwidth}
\begin{tabular}{c|ccccc|ccccc}
\toprule
\multirow{2}{*}{Personality} & \multicolumn{5}{c|}{LoRA (\%)}   & \multicolumn{5}{c}{IA3 (\%)}          \\ 
\cmidrule{2-11}
 & E/I & S/N & T/F & J/P & \checkmark/\xmark & E/I & S/N & T/F & J/P & \checkmark/\xmark \\ \midrule
ENFJ        & 54    & 51      & 52     & 60   & \checkmark & 54      & 52        & 68     & 71     & \checkmark \\
ENFP        & 56    & 63      & 64     & 67   & \checkmark & 63      & 58        & 56     & 56     & \checkmark \\
ENTJ        & 53    & 51      & 61     & 64   & \checkmark & 59      & 54        & 53     & 58     & \checkmark \\
ENTP        & 63    & 58      & 53     & 67   & \checkmark & 66      & 66        & 58     & \textcolor{red}{31}     & \xmark \\
ESFJ        & 79    & 51   & 52     & 51   & \checkmark & 56      & 57     & 56     & 67     & \checkmark \\
ESFP        & 56    & 54   & 73     & 67   & \checkmark & 54      & 52     & 53     & 51     & \checkmark \\
ESTJ        & 64    & 69   & 53     & 67   & \checkmark & 63      & 64     & 57     & 67     & \checkmark \\
ESTP        & 53    & 52   & 61     & 83   & \checkmark & 53      & 57     & 54     & 51     & \checkmark \\
INFJ        & \textcolor{red}{40}    & 59   & 66     & 67     & \xmark & 53       & 51        & 56     & 64     & \checkmark \\
INFP        & \textcolor{red}{44}    & 70   & 54     & 63     & \xmark & 72       & \textcolor{red}{41}        & 52     & 69     & \xmark \\
INTJ        & 57    & \textcolor{red}{45}     & 71     & 64     & \xmark & 58       & 51        & 52     & 65     & \checkmark \\
INTP        & 64    & 55        & 59     & 65     & \checkmark & 65       & 51        & 64     & 60     & \checkmark \\
ISFJ        & 53    & 66     & 56     & 71     & \checkmark & 53       & 63     & 51     & 58     & \checkmark \\
ISFP        & 53    & 60     & 60     & 64     & \checkmark & 69       & 66     & \textcolor{red}{43}     & 67     & \xmark \\
ISTJ        & 66    & 78     & 53     & 67     & \checkmark & 57       & 64     & 76     & 57     & \checkmark \\
ISTP        & 69    & 58     & 55     & 76     & \checkmark & 64   & 63     & 58     & 71     & \checkmark \\
\bottomrule
\end{tabular}
\end{adjustbox}
\caption{MBTI Personality Results. The scores at a given row correspond to the traits of that personality (ENTJ row scores are for E, N, T and J traits respectively). Note that \checkmark/\xmark  
\: indicates whether the composed PEM correctly aligns with the MBTI personality it was composed for or not.}
\label{tab:person}
\end{table}

The results for zero-shot classification are summarized in Table \ref{tab:trait} and \ref{tab:person} for individual trait PEMs and personality PEMs, respectively. In Table \ref{tab:trait}, we see that compared against a baseline BERT not fine-tuned via PEFT, the PEMs obtained for LoRA and IA3, both, outperform the baseline. Further, the trait PEMs correctly align with the trait they were trained for (seen by a score $>$ 50).

For the personality PEMs, as seen in Table \ref{tab:person}, the composed PEMs, correctly align with the personalities that they were composed for in most cases. Instances of mis-alignment are few and appear due to at most one misaligned trait. 

The output generated by the baseline and the PEM-equipped text generation model as an extrovert is listed as follows:

\begin{tcolorbox}[colback=green!10!white,colframe=green!50!black]

\emph{Prompt}: "Describe about your personal relationships." \\

\emph{Baseline model response}: "Write a list of 5 specific and specific ways you encourage and support your family members, friends, colleagues, or business associates ..." \\

\emph{PEM-equipped model response}: "I thrive on connecting with others and building meaningful bonds ..."

\end{tcolorbox}

\paragraph{Demo} We further create an API using Streamlit \footnote{API Demo Video: https://tinyurl.com/csci544-group-37} that allows a user to ask questions to our PEMs and interact with them. The API allows both question-answer style interaction as well as chatbot style conversational interaction.

\section{Conclusion and Future Work}
From our results, we conclude that PEFT and PEMs are effective domain learners and they can be merged for distribution generalization without additional fine-tuning to represent specialized composites for composite domains. 

For future work, additional compositions on PEMs can be explored that allow PEMs to unlearn, transfer domain expertise, multi-task and detoxify toxic language models. Trait and personality PEMs can be human evaluated and tested over a population to validate the findings of this work.

\clearpage
% \section{Division of Labor}

% \paragraph{Mann Patel} Ideated the project and led the group discussions, automated the evaluation process via a Selenium pipeline and formalized the algorithm, wrote adapter training pipeline, and ran experiments for composition associated with MBTI, worked on project presentation.

% \paragraph{Divyajyoti Panda} Wrote the evaluation script, performed PEM composition for IA3 through a script to select optimal $\{\lambda_i\}$ for each (personality) PEM, generated the results in Table \ref{tab:trait} and \ref{tab:person} for IA3, generated synthetic trait datasets.

% \paragraph{Hilay Mehta} Wrote the Streamlit API pipeline to allow for users to interact with the trait and personality PEMs, generated synthetic trait datasets using few shot prompting, and setting up the pipeline for merging \& training the PEMs.

% \paragraph{Parth Patel} Performed PEM composition for LoRA through a script to select optimal $\{\lambda_i\}$ for each (personality) PEM, generated the results in Table \ref{tab:trait} and \ref{tab:person} for LoRA, assisted with debugging the Streamlit API pipeline, generated synthetic trait datasets, worked on project presentation. 

% \paragraph{Dhruv Parikh} Wrote the training script for training LoRA and IA3 PEMs, helped with debugging text classification pipeline for training the individual PEMs, prepared the project presentation slides and wrote the project final report.

\clearpage
\bibliography{custom}

\end{document}